\documentclass[acmtog]{acmart}

\acmSubmissionID{980}

\usepackage{graphicx}
\usepackage{subfig}
\usepackage{array}    
\usepackage{booktabs} 
\usepackage{pifont}

\usepackage{hyperref}

\setcopyright{none}
\renewcommand\footnotetextcopyrightpermission[1]{}

\begin{document}


\begin{teaserfigure}
\centering
  \includegraphics[width=0.96\linewidth]{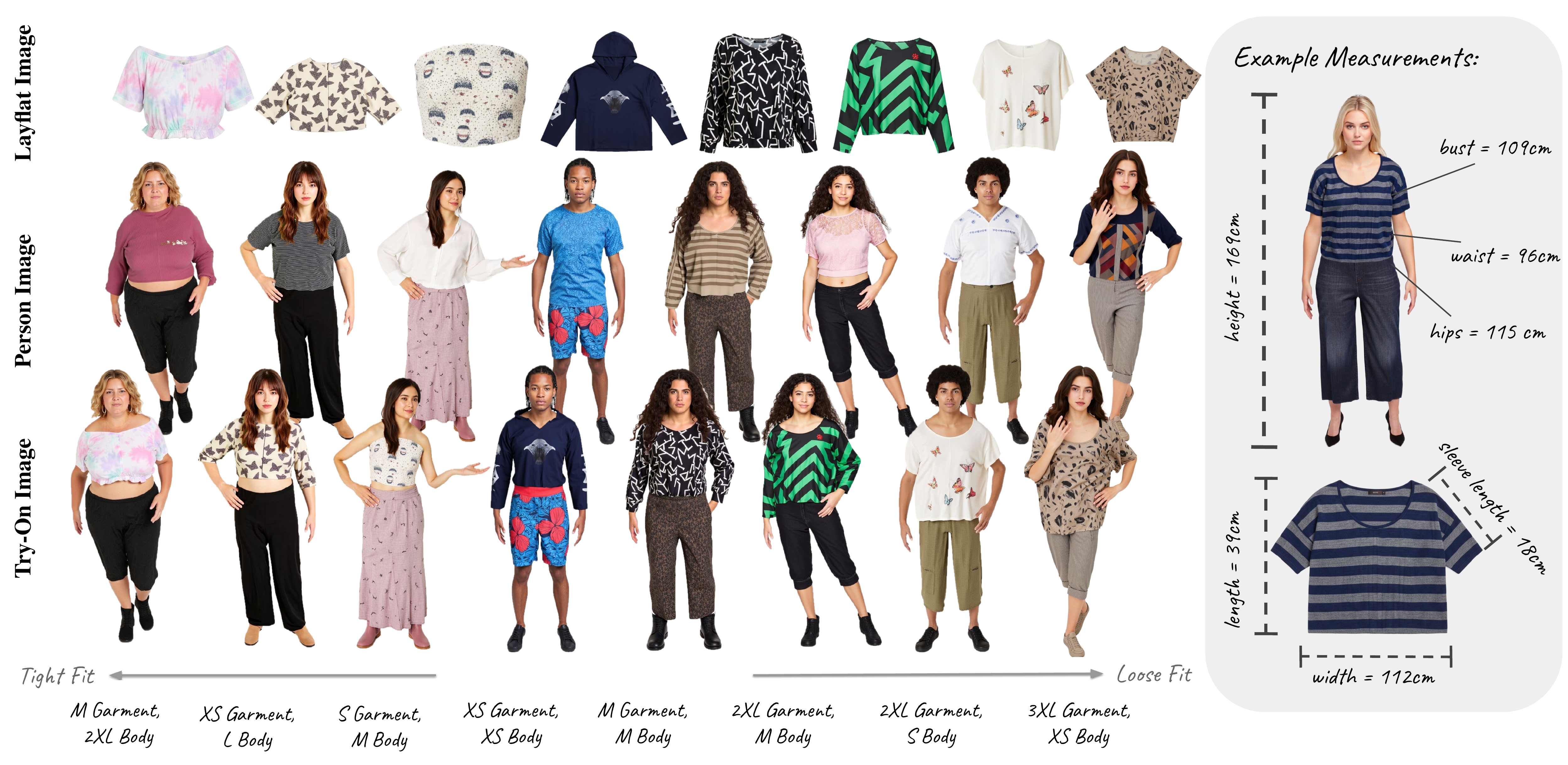}
  \Description{Teaser.}
  \caption{\textbf{The FIT Dataset.} We present FIT, a dataset and benchmark designed for \textit{fit-aware} virtual try-on, featuring diverse garment fits (e.g., tight, loose) and precise size annotations. \textit{Left:} Sample dataset triplets showing the conditioning garment image (\textit{top}), the conditioning person image (\textit{middle}), and the target try-on image (\textit{bottom}). \textit{Right:} Visualization of the corresponding person and garment measurement annotations. Backgrounds are removed for clarity.}
  \label{fig:teaser}
\end{teaserfigure}


\author{Johanna Karras}
\authornote{Equal contribution. Work done while authors were student researchers at Google.}
\affiliation{%
  \department{Paul G. Allen School for Computer Science and Engineering}
  \institution{University of Washington}
  \city{Seattle}
  \state{WA}
  \country{USA}}
\affiliation{%
  \department{}
  \institution{Google Research}
  \city{Seattle}
  \state{CA}
  \country{USA}}
\email{jskarras@cs.washington.edu}

\author{Yuanhao Wang}
\authornotemark[1]
\affiliation{%
  \department{Paul G. Allen School for Computer Science and Engineering}
  \institution{University of Washington}
  \city{Seattle}
  \state{WA}
  \country{USA}}
\affiliation{%
  \department{}
  \institution{Google Research}
  \city{Seattle}
  \state{CA}
  \country{USA}}
\email{yuanhaowang@cs.washington.edu}

\author{Yingwei Li}
\affiliation{%
  \department{}
  \institution{Google Research}
  \city{Mountain View}
  \state{CA}
  \country{USA}}
\email{yingweili@google.com}

\author{Ira Kemelmacher-Shlizerman}
\affiliation{%
  \department{Paul G. Allen School for Computer Science and Engineering}
  \institution{University of Washington}
  \city{Seattle}
  \state{WA}
  \country{USA}}
\affiliation{%
  \department{}
  \institution{Google Research}
  \city{Seattle}
  \state{CA}
  \country{USA}}
\email{kemelmi@google.com}



\title{FIT: A Large-Scale Dataset for Fit-Aware Virtual Try-On}







\begin{abstract}
Given a person and a garment image, virtual try-on (VTO) aims to synthesize a realistic image of the person wearing the garment, while preserving their original pose and identity. Although recent VTO methods excel at visualizing garment appearance, they largely overlook a crucial aspect of the try-on experience: the accuracy of garment fit -- for example, depicting how an extra-large shirt looks on an extra-small person.  A key obstacle is the absence of datasets that provide precise garment and body size information, particularly for ``ill-fit” cases, where garments are significantly too large or too small. Consequently, current VTO methods default to generating well-fitted results regardless of the garment or person size. 

In this paper, we take the first steps towards solving this open problem. We introduce FIT (\textbf{F}it-\textbf{I}nclusive \textbf{T}ry-on), a large-scale VTO dataset comprising over \textbf{1.13M} try-on image triplets accompanied by precise body and garment measurements. We overcome the challenges of data collection via a scalable synthetic strategy: (1) We programmatically generate 3D garments using GarmentCode~\cite{GarmentCode} and drape them via physics simulation to capture realistic garment fit. (2) We employ a novel re-texturing framework to transform synthetic renderings into photorealistic images while strictly preserving geometry. (3) We introduce person identity preservation into our re-texturing model to generate paired person images (same person, different garments) for supervised training. Finally, we leverage our FIT dataset to train a baseline fit-aware virtual try-on model. Our data and results set the new state-of-the-art for fit-aware virtual try-on, as well as offer a robust benchmark for future research. We will make all data and code publicly available on our \href{https://johannakarras.github.io/FIT/}{project page}: https://johannakarras.github.io/FIT.

\end{abstract}

\maketitle

\section{Introduction}
\label{sec:intro}


The rising popularity of online shopping and social media has increased the demand for virtual try-on (VTO) systems. Driven by advances in generative models, recent VTO works \cite{any2anytryon, catvton, mixmatch} have achieved remarkable progress in synthesizing photorealistic try-on images. However, they often merely transfer garment \textit{appearance} onto a person, neglecting to take into account the person or garment sizes. As such, current VTO methods fail to address a fundamental question for any user: \textit{"How will this garment actually \text{fit} me?"} 
This severely limits the accuracy and reliability of  existing VTO tools to simulate a real-life try-on experience. Furthermore, it prevents users from experimenting with different sizes to achieve a desired fitted or oversized look. Consequently, there is significant commercial and research interest in developing a fit-aware VTO method.

Fit-aware try-on remains challenging due to the scarcity of real-world data annotated with precise person and garment measurements. Most existing VTO datasets \cite{viton-hd,deepfashion,deepfashion2,viton,cloth3d,cloth4d,tailornet,dresscode,sewformer,deepfashion3d,streettryon} are curated by scraping catalog images from online retailers, which inherently lack "ill-fit" examples, i.e. the garment is too large or too small. Moreover, while some retailers provide size metadata, these annotations are often non structured and difficult to process at scale. Synthetic 3D garments created by artists offer an alternative, but this data suffers from limited scale and realism.



To fill this gap, we introduce FIT (\textbf{F}it-\textbf{I}nclusive \textbf{T}ry-on), the first large-scale, size-aware VTO benchmark explicitly designed to capture diverse upper-garment fit scenarios. By pivoting to a synthetic data generation pipeline (GarmentCode~\cite{GarmentCode}), we overcome the limitations of real-world data collection. We procedurally create 3D garments with exact ground-truth measurements and simulate their drape onto a wide range of parametric bodies. This approach ensures not only size measurements, but also details like wrinkles, stretch, and garment coverage, are physically accurate. 
To close the domain gap between synthetic and real images, we employ a novel re-texturing pipeline designed to generate photorealistic textures for the synthetic renderings, while ensuring that the garment fit and body shape are preserved. To this end, we fine-tune a foundational image generation model, Flux.1-dev\cite{flux1dev}, to generate realistic person images from the synthetic normal maps and text-based garment descriptions.

Another critical bottleneck in VTO research is the lack of paired training data (identical subject and pose, different garments). Consequently, existing methods \cite{mixmatch, tryondiffusion, catvton, kim2025promptdresser,xu2025ootdiffusion,kim2024stableviton} are forced to formulate VTO as a self-supervised reconstruction task, which limits real-world applications, or rely on synthesized pseudo triplets \cite{any2anytryon,du2023greatness,zhang2025boow}, which suffer from inaccurate masking, identity loss, and size leakage.  In contrast, our synthetic pipeline offers the unique advantage of controllability. We can simulate the same 3D subject in the same pose wearing multiple distinct garments, thereby generating ground-truth paired person data. Building on this insight, we further propose a novel framework for paired person image generation that ensures accurate 3D grounding and identity preservation.

Our dataset contains 1.13M  training and 1K test samples of both men's and women's upper-garments. Each sample consists of a target try-on image, layflat garment image, a paired person image, as well as person and garment measurements. Our target try-on images cover diverse fit scenarios, including extreme ill-fits (e.g. a size 3XL draped onto a size XS person). By fine-tuning Flux.1-dev\cite{flux1dev} with our custom dataset and a custom measurement encoder, we demonstrate a baseline fit-aware VTO model that accurately showcases garment fit.

To summarize, we present the following contributions:
\begin{enumerate}
\item We introduce \textbf{FIT}, the first large-scale dataset and benchmark explicitly designed for fit-aware virtual try-on, featuring precise metric annotations and diverse fit scenarios.
\item We develop a scalable synthetic data generation pipeline that leverages physics simulation and generative re-texturing to produce photorealistic try-on triplets with 3D grounding.
\item We demonstrate a novel, fit-aware virtual try-on model (\textbf{Fit-VTO}) that incorporates person and garment measurements to visualize not only garment appearance, but also accurate garment fit.
\end{enumerate}
\section{Related Works}
\label{sec:related-works}

\begin{figure*}[!tbp]
  \centering
  \vspace{-3mm}
  \includegraphics[width=\linewidth]{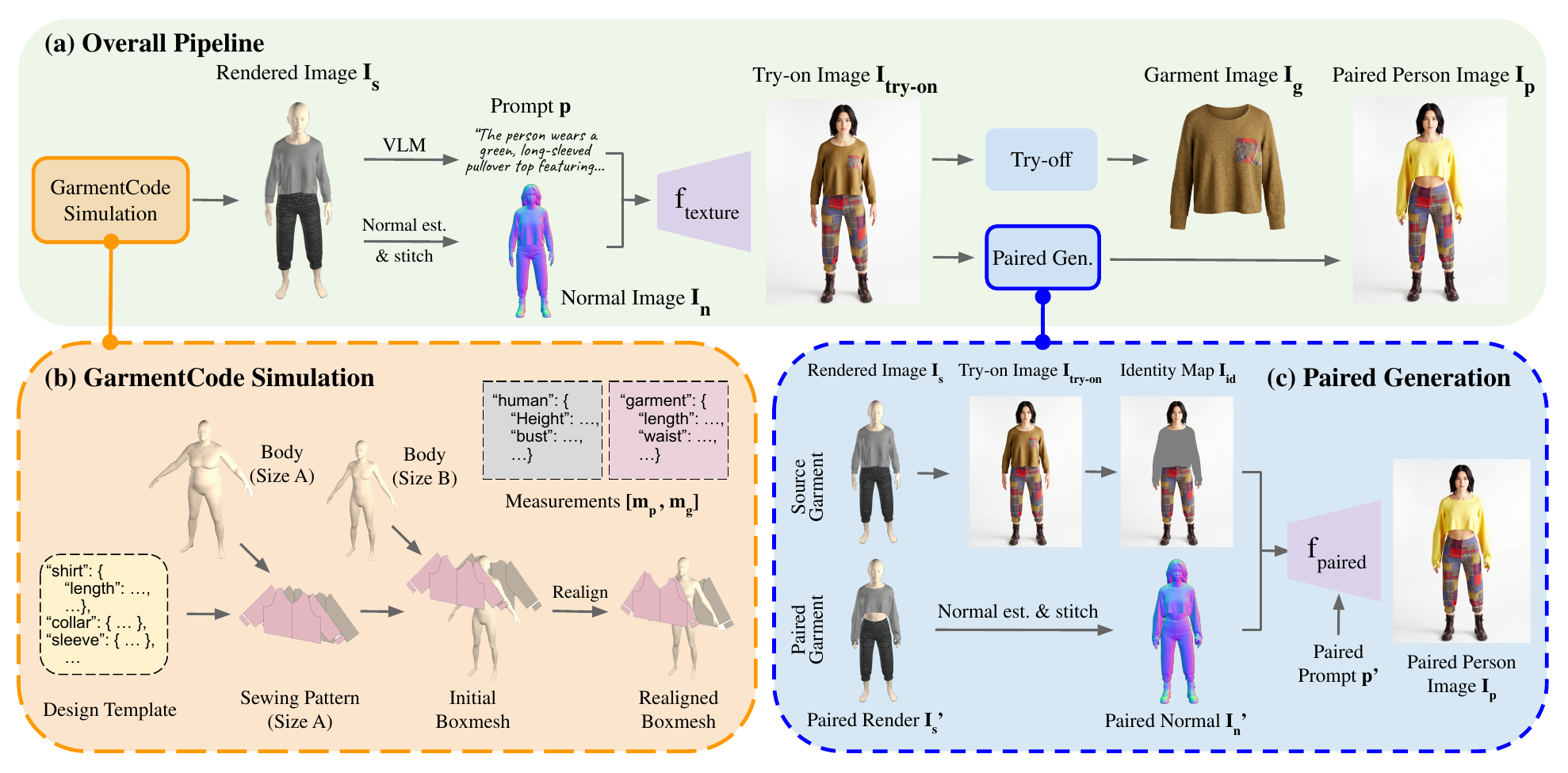}
  \vspace{-5mm}
  \Description{FIT data generation pipeline.}
  \caption{\textbf{FIT dataset generation.} 
  \textbf{(a)} Overall pipeline: For each sample, we first simulate garment draping in 3D via GarmentCode, rendering a synthetic try-on image $I_s$ (see (b)). Then, we generate a text prompt $p$ (via VLM) describing the person and garment appearance, as well as a composite normal map $I_n$ based on $I_s$. We use $p$ and $I_n$ to condition our re-texturing model $f_\text{texture}$ to generate the photorealistic try-on image $I_{\text{try-on}}$. Finally, our model $f_\text{paired}$ generates a paired person image $I_p$ (see (c)) and a VLM synthesizes the corresponding layflat garment $I_g$.
  \textbf{(b)} GarmentCode simulation: Given a garment design template, we compute a sewing pattern with measurements $m_g$ for a specific body size $A$. Then, we cross-drape the pattern onto a different target body of size $B$ with person measurements $m_p$, using box-mesh realignment to prevent simulation failures. \textbf{(c)} To generate a paired person image (same person and pose, different garment), we start with a paired rendered image $I_s'$ containing a \textit{different} garment than in $I_\text{try-on}$ draped onto the \textit{same} body. Next, we derive an identity map $I_\text{id}$ by masking out the combined source and paired garment regions in $I_{\text{try-on}}$. Conditioned on $I_\text{id}$, the paired normal map $I_n'$, and a paired prompt $p'$, $f_\text{paired}$ generates $I_p$.}
  \label{fig:data_generation_pipeline}
  \vspace{-3mm}
\end{figure*}

\subsection{Virtual Try-On Datasets}

A primary bottleneck for fit-aware virtual try-on is the lack of datasets containing explicit size annotations or ill-fitting examples. Standard 2D benchmarks, such as ViTON~\cite{viton}, ViTON-HD~\cite{viton-hd}, DressCode~\cite{dresscode}, StreetTryOn~\cite{streettryon}, and LAION-Garment~\cite{any2anytryon}, predominantly feature well-fitted garments, lacking the diverse fit conditions required for size-aware training. While some datasets, including SIZER~\cite{sizer}, SV-VTO~\cite{yamashita2024size}, and Fit4Men~\cite{fitcontroler} collect real-world samples for this purpose, they remain limited in scale and diversity. See Table~\ref{tab:dataset_comparison}.


Alternatively, 3D datasets~\cite{cloth3d, deepfashion3d, cloth4d, sewformer, sizer} offer 3D models of clothed humans. 
However, extracting accurate garment measurements from raw meshes is often infeasible. GarmentCode~\cite{GarmentCode} addresses this by introducing a domain-specific language for generating sewing patterns with explicit size parameters, enabling synthetic garment generation across varied garment and body sizes~\cite{GarmentCodeData}. However, for extreme ill-fitting garment draping cases, GarmentCode tends to produce significant and frequent draping errors. Furthermore, raw 3D synthetic datasets~\cite{GarmentCodeData, garmagenet} generally suffer from their lack of realistic textures, which leads to poor real-world generalization. Although Sewformer~\cite{sewformer} attempts to enhance realism via texture synthesis and SDEdit refinement, the results are still cartoonish and lack fit diversity. To bridge these gaps, we adapt GarmentCode for ill-fit scenarios, as well as introduce a novel pipeline for transforming synthetic GarmentCode renderings into photorealistic images. 

\begin{table}[htbp]
    \centering
    \small
    \caption{Comparison of related datasets. We compare FIT to several related datasets. For scale, we report the number of training images.}
    \vspace{-8pt}
    \label{tab:dataset_comparison}
    \begin{tabular}{|l|c|c|c|c|c|}
        \hline
        Dataset& Realism & Ill-Fit & Measurements & Triplet & Scale \\
        \hline
        SV-VTO & \checkmark & \checkmark & \checkmark & \checkmark & 1,524\\ 
        SIZER & \checkmark & \checkmark & \checkmark & \ding{55} & 2,000 \\
        \small{DeepFashion3D}& \ding{55} & \ding{55} & \ding{55} & \ding{55} & 2,078 \\ 
        ViTON-HD & \checkmark & \ding{55} & \ding{55} & \ding{55} & 11,647 \\ 
        Size4Men & \checkmark & \checkmark & \ding{55} & \ding{55} & 13,000 \\ 
        LAION-Garment& \checkmark & \ding{55} & \ding{55} & \checkmark & 60K\\ 
        SewFactory & \checkmark  &  \ding{55} & \checkmark & \ding{55} & 1M \\ 
        \small{GCD} & \ding{55} & \ding{55} & \checkmark & \ding{55} & 115K\\ 
        \textbf{Ours} & \checkmark & \checkmark & \checkmark & \checkmark & 1.13M \\
        \hline
    \end{tabular}
    \vspace{-12pt}
\end{table}

\subsection{Image-Based Virtual Try-On}

Image-based virtual try-on methods are generally categorized into two paradigms: \textit{mask-based}, which utilize explicit segmentation maps to localize generation, and \textit{mask-free}, which synthesize results directly without segmentation priors.

\paragraph{Mask-Based Methods} These approaches formulate virtual try-on as a conditional inpainting task, where the target clothing region is masked and filled based on the garment image and human priors. Early warping-based works~\cite{viton, viton-hd} established a two-stage paradigm: warping the garment to the target body followed by refinement. Recent approaches have shifted toward single-stage diffusion-based architectures, achieving state-of-the-art photorealism~\cite{tryondiffusion, streettryon, mixmatch, catvton, xu2025ootdiffusion, kim2024stableviton}. However, because these methods rely on inpainting within a fixed mask, they primarily focus on texture preservation and body alignment, largely neglecting the physical reality of garment sizing.

\paragraph{Mask-Free Methods.}
Another line of research \cite{issenhuth2020not,ge2021disentangled,ge2021parser,dumitigating,du2023greatness,zhang2025boow,any2anytryon} focus on mask-free architectures. Since real-world paired data is unavailable, these methods typically rely on generating ``pseudo-triplets'' via generative modeling to enable supervised training. A common strategy involves a ``Teacher-Student'' distillation framework, where a mask-based ``teacher'' model swaps garments on training images to generate synthetic ground-truth for a mask-free ``student''. Similarly, Any2AnyTryOn~\cite{any2anytryon} leverages a pre-trained inpainting model to digitally replace garments in the try-on region. A fundamental bottleneck is that this training data is itself hallucinated, causing models to inherit the artifacts and geometric inconsistencies of the teacher. In contrast, our synthetic pipeline simulates actual draping dynamics on 3D bodies, yielding true ground-truth pairs with precise geometry and segmentation, effectively bypassing the error accumulation of 2D pseudo-triplet generation.

\paragraph{Fit and Size Control.}
While most VTO works ignore size, a few attempts have been made to incorporate fit information using geometric heuristics~\cite{chen2023size, fitcontroler, kuribayashi2023image, yamashita2024size}. For instance, \cite{chen2023size} leverages clothing landmarks to transform garment size, while \cite{kuribayashi2023image} uses body-to-clothing ratios to resize the conditioning segmentation maps. More recently, \cite{yamashita2024size} and \cite{fitcontroler} introduce coarse fit conditioning based on descriptors (e.g., ``tight" or ``loose''). However, by relying on imprecise intermediate values or coarse labels, past methods struggle to generalize to complex poses and lack precise control. In contrast, our fit-aware model avoids noisy geometric heuristics by conditioning on exact metric measurements.

\section{Fit-Inclusive Try-on (FIT) Dataset}
\label{sec:dataset}

\begin{figure*}[!tbp]
  \centering
  \vspace{-3mm}
  \includegraphics[width=1.0\linewidth]{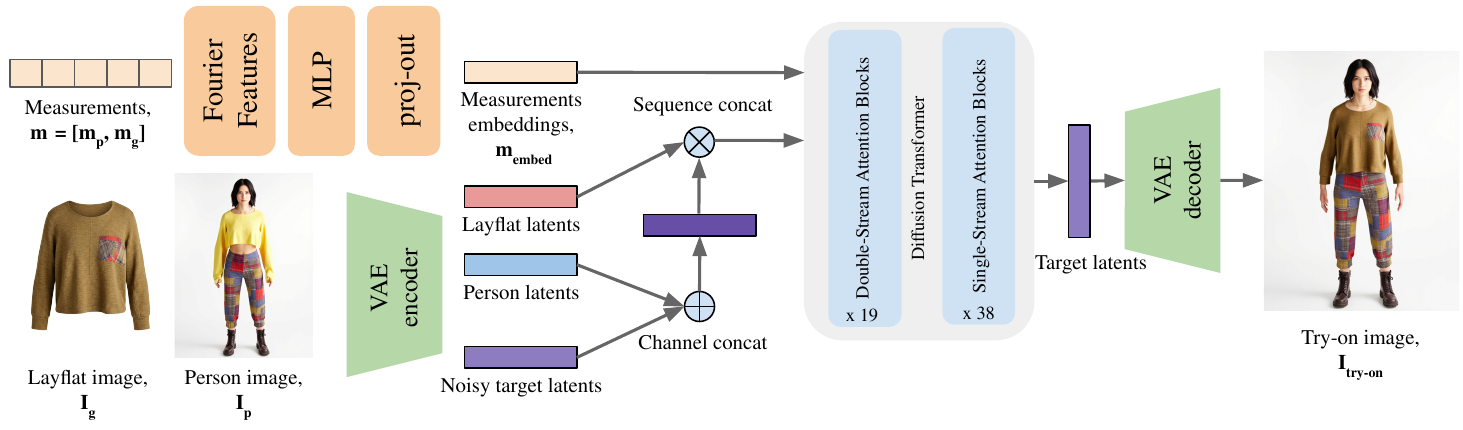}
  \Description{Fit-VTO architecture.}
  \caption{Fit-VTO architecture. Our architecture is a flow-based diffusion model based on Flux.1-dev~\cite{flux1dev} and finetuned with LoRA~\cite{lora}. FiT-VTO generates a try-on image $I_{\text{try-on}}$ given a layflat garment image $I_g$, paired person image $I_p$, and person-garment measurements  $m = [m_p, m_g]$. First, image inputs $I_g$ and $I_p$ are encoded into latents separately through a pre-trained VAE encoder. We replace the text embeddings in Flux.1-dev with custom measurement embeddings $m_{\text{embed}}$ computed from $m$. Person latents are channel-concatenated with the noisy target latents, while layflat latents and  $m_{\text{embed}}$ are sequence-wise concatenated with $z_t$. After processing through the diffusion transformer, clean latents are decoded by the VAE decoder.}
  \label{fig:architecture}
  \vspace{-3mm}
\end{figure*}

In this section, we describe the construction of the FIT dataset. We first report the dataset statistics in Section~\ref{data-stats}. We then detail our data generation pipeline, illustrated in Figure~\ref{fig:data_generation_pipeline}, which consists of the following steps: (1) procedurally generating garment assets with measurements $m_g$ and simulating their drape across diverse sizes of bodies with measurements $m_p$ via GarmentCode (Section~\ref{ssec:data-garmentcode}); (2) transforming the synthetic renderings $I_s$ into photorealistic try-on images $I_{\text{try-on}}$ via a geometry-preserving re-texturing framework (Section~\ref{ssec:data-sim2real}); (3) leveraging identity conditioning to generate a paired person reference image $I_{p}$ featuring the same person wearing a different garment (Section~\ref{ssec:data-triplet}); and (4) synthesizing the corresponding layflat garment image $I_g$ using an off-the-shelf VLM model~\cite{pictor} (Section~\ref{ssec:data-tryoff}).

\subsection{Dataset Statistics} \label{data-stats}
Our dataset consists of 1,137,282 training and 1000 test samples, each consisting of $(I_{\text{try-on}}, I_{\text{p}}, I_g, m_p, m_g)$. Our data covers 168 distinct body shapes (82 men's, 86 women's) in sizes XS-3XL, 528 body poses,
as well as 158,483 unique top and garment designs. Our dataset covers a diverse range of fits, from loose to tight fits. We provide a histogram of each person/garment size combination in the appendix. The test dataset is balanced to match the overall distribution over gender, body sizes, and person/garment size combinations.  See Figure~\ref{fig:teaser} and the appendix for examples of our dataset. 




\subsection{GarmentCode Simulation} \label{ssec:data-garmentcode}
GarmentCode \cite{GarmentCode} is a parametric programming framework that enables the procedural generation and draping of 3D garment patterns, allowing for precise control over sizing and design details. 

To generate try-on images with diverse fits, we implement a cross-draping strategy. We begin by sampling various garment templates and human body models with known measurements $m_p$. From a garment template, we generate sewing patterns fitted to multiple human bodies of varying sizes. We then simulate draping these sewing patterns onto a single target human model via GarmentCode's custom implementation of Warp~\cite{macklin2022warp}, thereby creating realistic "tight" and "loose" fit scenarios. 
However, direct cross-draping initially fails because the 3D box-mesh specified by standard sewing patterns is aligned with its original target body, causing severe misalignments when applied to a new body. We address this by explicitly realigning the initial box-mesh panels to the target mesh position before simulation. Please refer to the appendix for details. Furthermore, GarmentCode's default implementation stitches top and bottom garments together into a unified mesh, preventing the appearance of ``tucked-out" shirts. We modify this behavior to drape the top and bottom garments in two separate steps (typically simulating the bottom garment first) to ensure proper layering and realistic interactions between items. The draped 3d mesh is then reposed and rendered with different person poses to form a synthetic rendering image $I_s$.


Our procedural framework allows us to programmatically extract precise ground-truth garment measurements $m_g$ in centimeters directly from the 2D sewing pattern specifications. We focus on five critical metrics used in standard sizing: garment length (high point shoulder to hem), bust circumference (width), and sleeve length for tops; and waist and out-seam length for bottoms. We also derive four key body measurements directly from GarmentCode's parametric body model: height, bust, waist, and hips.

\subsection{Synthetic-to-Photorealistic Retexturing} \label{ssec:data-sim2real}

Our re-texturing pipeline is designed to transform the synthetic rendering $I_s$ into a photorealistic image while strictly preserving the geometry of both the garment and the subject. Due to the lack of paired synthetic-to-real training data, we utilize surface normal maps as a geometry-preserving bridge between domains. Specifically, we fine-tune a diffusion model, $f_{\text{texture}}$ (based on Flux.1-dev~\cite{flux1dev}), to synthesize photorealistic textures conditioned on an input normal map $I_n$ and a text prompt $p$. 
$f_\text{texture}$ is trained on real-world images with the following objective:
\begin{equation}
    \hat{I}_{\text{try-on}} = f_{\text{texture}}(I_n, p)
\end{equation}
where $I_n = N(I_{\text{try-on}})$ represents the normal map extracted from an off-the-shelf estimator $N$~\cite{khirodkar2024sapiens}. 

Despite utilizing normal maps, a significant domain gap persists between real-world and synthetic data. First, synthetic renderings $I_s$ lack anatomical details, featuring bald heads and bare feet. To address this, we employ a composite refinement strategy: we prompt Nano Banana Pro~\cite{nanobananapro} to inpaint realistic facial features, hair, and footwear onto $I_s$, estimate the normals of this enhanced image, and stitch the resulting head and feet regions onto the original synthetic normal map. This ensures realistic semantic cues while leaving the body and garment geometry untouched. Second, Similarly, synthetic meshes lack intricate surface details, such as pockets, buttons, and seams. We observe that our model $f_\text{texture}$ successfully inpaints these details when guided by appropriate text prompts. Similarly, due to GarmentCode's limited controllability of material, the synthetic garments exhibit uniform, smooth fabric. To increase fabric diversity, we sample from 72 fabric types (e.g. leather, cotton, silk) and inject it into the text prompt. We further align the domains by augmenting the training data with random normal map blurring. This simulates the smoothness of synthetic normal maps and improves generation quality.

\subsection{Paired Person Reference Image Generation} \label{ssec:data-triplet}

Our synthetic framework enables the generation of ground-truth paired data by exploiting procedural controllability. By fixing the subject's shape and pose while draping two distinct garments, we obtain pairs of synthetic renderings $(I_s, I_s')$, normal maps $(I_n, I_n')$, garment masks $(m_g, m_g')$, and prompts $(p, p')$.

First, we generate the primary try-on image $I_{\text{try-on}} = f_{\text{texture}}(I_n, p)$ using the re-texturing pipeline described in Section~\ref{ssec:data-sim2real}. Next, to synthesize the paired reference image $I_p$, we employ a conditional inpainting model $f_{\text{paired}}$:
\begin{equation} \label{eq:sim2real}
    I_p = f_{\text{paired}}(I_\text{id}, I_n', p'),
\end{equation}
where $I_{id}$ represents the identity map, defined as $I_{id} = I_{\text{try-on}} \odot (\neg m_g \cap \neg m_g')$. This operation preserves the skin and background from the try-on image while masking out the regions occupied by both the source and paired garments. Essentially, $f_{\text{paired}}$ serves as a geometry-guided inpainter. To train $f_{\text{paired}}$, we utilize real human images, creates identity maps by estimating garment masks and applying random dilation to mimic the dual-garment masking seen at inference. Additionally, we limit our scope to upper-body try-on, hence we enforce identical bottom garment geometry across pairs during simulation. In practice, we train a unified model for $f_\text{texture}$ and $f_\text{paired}$ following Eq. \ref{eq:sim2real}, but randomly dropping out $I_\text{id}$.

\subsection{Layflat Image Generation} \label{ssec:data-tryoff}

Motivated by the impressive image synthesis capability of Nano Banana Pro~\cite{nanobananapro}, we use it as an off-the-shelf virtual try-off model to generate a layflat garment image $I_g$ from $I_\text{try-on}$. Please refer to the appendix for the exact prompts used.

\section{Fit-Aware Virtual Try-On}
\label{sec:vto}

Given an image $I_p$ of person $p$, a garment image $I_g$ of target garment $g$, target garment measurements $m_g$, and person measurements $m_p$, our Fit-VTO model $f_{\text{vto}}$ synthesizes the predicted try-on result $\hat{I}_{\text{try-on}}$ of person $p$ wearing $g$ according to the measurements $m_p$ and $m_g$.

\begin{equation}
    \hat{I}_{\text{try-on}} = f_{\text{vto}}(I_p, I_g, m_p, m_g)
\end{equation}

\subsection{Dataset Preparation}\label{ssec:dataset-preparation}
To increase the robustness of our model to diverse, real-world garments and poses, we crawled 330,559 online fashion images and their corresponding layflat garment images $I_g$, to augment our FIT training dataset. 
Since ground-truth measurements are not available for online images, we set the measurements to null values ($-1$). For FIT data samples, all measurements $m_p$, $m_g$ are normalized between 0 and 1.

\subsection{Architecture}

Our architecture (Figure~\ref{fig:architecture}) is a flow-matching diffusion model $x_{\theta}$ represented as:

\begin{equation}
    \hat{v}_{t} = x_{\theta}(z_t, t, I_p, I_g, m_p, m_g)
\end{equation}

where $z_t$ is the noisy ground-truth image $x_0$ at diffusion timestep $t$ and $\hat{v}_{t}$ is the predicted velocity. The model $x_\theta$ is trained to satisfy the consistency constraint where $\hat{v}_{t}$ approximates ground-truth velocity $v_t = x_0 - z_0$. 

Our network $x_{\theta}$ is finetuned from the pre-trained Flux.1-dev text-to-image model~\cite{flux1dev}. FLUX.1-dev is a powerful, 12 billion parameter text-to-image generator that employs a rectified flow formulation and a Multi-modal Diffusion Transformer (MMDiT) backbone for efficient, high-fidelity image synthesis. 
We finetune only the lightweight LoRA parameters, keeping the majority of the original model weights frozen.

\noindent\textbf{Person and Garment Conditioning.} We condition the model on paired person image $I_p$ and garment image $I_g$. Since the $I_p$ is pixel-aligned to the noisy target image $z_t$, we concatenate latents from $I_p$ and $z_t$ channel-wise. Since $I_g$ latents need to be warped to $z_t$, we concatenate them along the sequence dimension after packing.

\noindent\textbf{Measurement Conditioning.} To condition on person and garment measurements, we remove the CLIP and T5 text conditionings for Flux.1-dev and instead condition with measurement embeddings from our custom measurement encoder $\mathcal{E}_m$. We first concatenate person measurements $m_p$ with garment measurements $m_g$ into a measurement vector $m = [m_p, m_g] \in R^7$. Then, we compute the Fourier Feature Embeddings for each measurement with 8 Fourier frequency bands, mapping $m \rightarrow m_{\text{embed}} \in R^{7 \times 16}$. These embeddings are further processed by an MLP and projected to the hidden dimension $R^{3072}$ of the MMDiT. Our model is conditioned on $m_{\text{embed}}$ with positional encodings for each measurement via cross-attention, replacing the T5 text conditioning in the single-stream and double-stream blocks.



\section{Experiments}
\label{sec:experiments}

We describe details of experiments in this section. We quantitatively and qualitatively evaluate the quality of our synthetic triplet data and demonstrate the effectiveness of our baseline fit-aware VTO model against state-of-the-art methods.


\subsection{Implementation Details}\label{ssec:implementation-details}
For synthetic data generation, we initialize our re-texturing model from the pre-trained Flux.1-dev~\cite{flux1dev} checkpoint and only finetune with LoRA layers~\cite{lora} with rank $64$ and alpha $64$. The model is trained on a custom dataset of 50k real person images (see appendix for details). We adopt Prodigy optimizer with learning rate $1.0$ and weight decay factor $0.01$. The training is done on 8 H200 GPUs with a total batch size of $64$ and 5k training iterations (~1 day).

Our baseline VTO model initialized from Flux.1-dev checkpoint and fine-tuned using LoRA layers~\cite{lora} with rank $128$ and alpha $128$. The measurement encoder is zero-initialized for stable early training. 
We fine-tune our model for 2M iterations on a mix of FIT training dataset and real-world images. 
The learning rate is $10^{-4}$ with $1000$ warm-up steps and batch size is $64$. All training is done on 64 TPU-v5's for ~2 days. At inference, we set guidance scale to 1.0 and number of inference steps is set to 50. We keep the same inference scheduler as the base Flux.1-dev release.

\subsection{Evaluation Baselines, Datasets \& Metrics}
\noindent \textbf{Paired Image Generation Evaluation.}
In this work, we propose a novel framework for generating pseudo-ground-truth paired-person images to enable mask-free VTO training. We benchmark against three baseline strategies: (1) \textbf{VLM-based}, which prompts Large Vision Language Models to swap garments while preserving context; (2) \textbf{VTO-based}, which utilizes off-the-shelf virtual try-on models for garment transfer; and (3) \textbf{Inpainting-based}, which replaces masked garment regions via generative inpainting. We implement these baselines using Nano Banana Pro, CatVTON~\cite{catvton}, and FLUX-Controlnet-Inpainting \cite{fluxinpaint}.

To quantify how well the paired image $I_p$ preserves the identity of the original $I_{\text{try-on}}$ in non-garment regions (i.e., background, head, and limbs), we compute the Masked L1 Distance $\mathcal{L}_{\text{id}}$:
\begin{equation}
\mathcal{L}_{\text{id}} = \frac{1}{N} \sum \left| (I_p - I_{\text{try-on}}) \odot M \right|,
\end{equation}
where $M = \mathbf{1} - (m_g \cup m_g')$ represents the binary mask of the non-garment regions, and $N$ is the number of valid pixels in $M$. We randomly sampled 1000 cases from the dataset to compute this metric. The pixel values are between 0 and 255.\\

\noindent \textbf{Fit-Aware VTO Evaluation.}
We compare our method qualitatively and quantitatively to Any2AnyTryon~\cite{any2anytryon}, Nano Banana Pro~\cite{nanobananapro}, COTTON~\cite{chen2023size}, IDM-VTON~\cite{idmvton}, and ablated versions of our method. We provide implementation details about related methods in the appendix. We evaluate on the VITON-HD test dataset~\cite{viton-hd} to measure general try-on accuracy and the FIT test dataset to evaluate fit-aware try-on accuracy. For VITON-HD, we generate paired-person images according to Section~\ref{ssec:data-triplet}.

We compute common VTO metrics -- SSIM~\cite{ssim}, FID~\cite{fid}, LPIPS~\cite{lpips}, KID~\cite{kid} -- to evaluate image similarity between ground-truth and synthesized try-on images. We also implement a custom metric (IoU), specifically designed for measuring size fidelity for the FIT dataset. IoU measures the Intersection-Over-Union of the garment mask in synthesized try-on image and the ground truth. We do not compute IoU for VITON-HD, as this dataset does not provide any size conditioning.

\begin{figure}[!tbp]
  \centering
  \includegraphics[width=\linewidth]{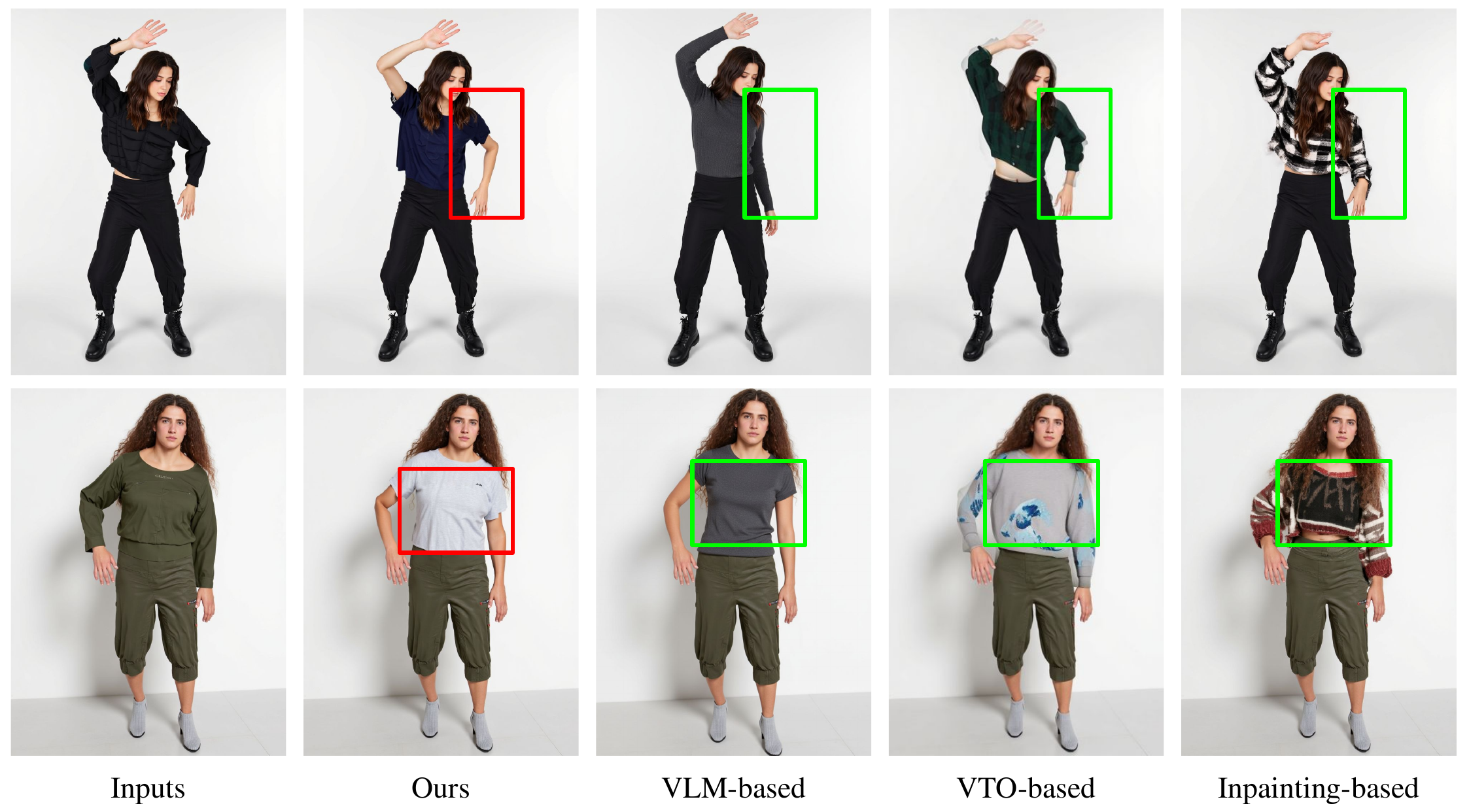}
  \Description{Paired Image Generation Comparison.}
  \vspace{-5mm}
  \caption{Paired Image Generation Comparison. VLM methods struggle with pose and shape preservation, while VTO and inpainting baselines introduce artifacts. Our approach yields highly consistent paired data.}
  \label{fig:triplet-compare}
  \vspace{-5mm}
\end{figure}

\subsection{Paired Image Evaluation Results}
We present a qualitative comparison of the generated paired-person images in Figure~\ref{fig:triplet-compare}. Despite their impressive editing capabilities, VLM-based methods fail to guarantee identity preservation in non-garment regions (e.g., the left arm pose deviation in the top row). Furthermore, they often disregard the underlying body shape within the garment region (e.g., the inconsistent chest volume in the bottom row). Similarly, the VTO and Inpainting-based baselines introduce significant visual artifacts and struggle to maintain geometric consistency. In contrast, our approach achieves near-perfect identity and body shape preservation by explicitly conditioning on the identity map and ground-truth normals.
Quantitative analysis confirms our visual findings: our method achieves an $\mathcal{L}_{\text{id}}$ of 1.61, significantly outperforming VLM-based (4.45), VTO-based (2.29), and Inpainting-based (3.91) baselines.
These results demonstrate that our pipeline successfully generates highly consistent paired data essential for robust VTO training.
\newcommand{\shortminus}{\scalebox{0.75}[1.0]{$-$}}
\begin{table*}[htbp]
    \centering
    \small
    \caption{Quantitative comparisons. We compare Fit-VTO to related methods and ablated versions of our method. Ours$_{\textbf{ft\_vitonhd}}$ refers to our method finetuned with VITON-HD training data. \textbf{Bolded} and \underline{underlined} values indicate the best and second-best scores per column, respectively.}
    \label{tab:comparisons}
    \vspace{-2mm}
    \begin{tabular}{|c|c|c|c|c||c|c|c|c|c|}
        \hline
        & \multicolumn{4}{c||}{ VITON-HD } & \multicolumn{5}{c|}{ FIT Dataset } \\
        \hline
        & \text{SSIM} $\uparrow$ & \text{FID} $\downarrow$ & \text{LPIPS} $\downarrow$  & \text{KID} $\downarrow$ & \text{SSIM} $\uparrow$ & \text{FID} $\downarrow$ & \text{LPIPS} $\downarrow$  & \text{KID} $\downarrow$ & \text{IOU} $\uparrow$ \\
        \hline
        Any2AnyTryon~\cite{any2anytryon} & 0.758 &  14.186 & 0.152 & 2.413 & 0.819 & 25.059 & 0.209 & 3.939 &  0.783 \\
        Nano Banana Pro~\cite{pictor} & 0.552 & 11.344 & 0.501 & \underline{0.624} &  0.785 & 19.926 & 0.166 & 1.676 & 0.792 \\
        COTTON~\cite{chen2023size} & 0.615 & 39.117  & 0.349 & 11.397 & 0.759 & 29.716 & 0.207 & 6.269 & 0.739 \\
        IDM-VTON~\cite{idmvton} & \textbf{0.849} & \textbf{9.115} & \textbf{0.077} & \textbf{0.471} & 0.739 & 31.229 & 0.246 & 6.819 & 0.789 \\
        Ours$_{\text{no FIT}}$ & 0.817 & 11.499 & 0.103 & 0.639 & 0.852 & 16.427 & 0.095 & 0.849 & 0.844 \\
        Ours$_{\text{text}}$ & 0.763 & 11.367 & 0.134 & 0.766 & 0.911 & 11.624 & 0.054 & 0.576 & 0.932 \\
        Ours$_{\text{FIT only}}$& 0.732 & 14.651 & 0.192 & 1.061 & \underline{0.912} & \underline{11.248} & \underline{0.052} &  \underline{0.532} & \underline{0.952} \\
        \textbf{Ours} & 0.817 & 11.391 & 0.102 & 0.651 & \textbf{0.914} & \textbf{10.381} & \textbf{0.050} & \textbf{0.144} & \textbf{0.955} \\
        \textbf{Ours}$_{\textbf{ft\_vitonhd}}$ & \underline{0.833} & \underline{9.320} & \underline{0.087} & 0.670 & 0.846 & 17.041 & 0.096 & 1.724 & 0.908 \\
        \hline
    \end{tabular}
    \vspace{-3mm}
\end{table*}

\subsection{Fit-VTO Qualitative Results}
We showcase qualitative results of Fit-VTO on the synthetic FIT test dataset in Figure~\ref{fig:qualitative_vto}. Our method synthesizes high-quality try-on images that maintain high fidelity to the person identity and garment appearance, while accurately reflecting realistic garment fit with respect to the person and garment measurements. Fit-VTO handles diverse fit cases, including tight fit, perfect fit, and loose fit. Our Fit-VTO method also generalizes to real-world images (VITON-HD~\cite{viton-hd}) without measurements, as shown in the bottom two rows of Figure~\ref{fig:comparison}. 

\begin{figure*}[!tbp]
  \centering
  \vspace{-0.8em}
  \includegraphics[width=0.9\linewidth]{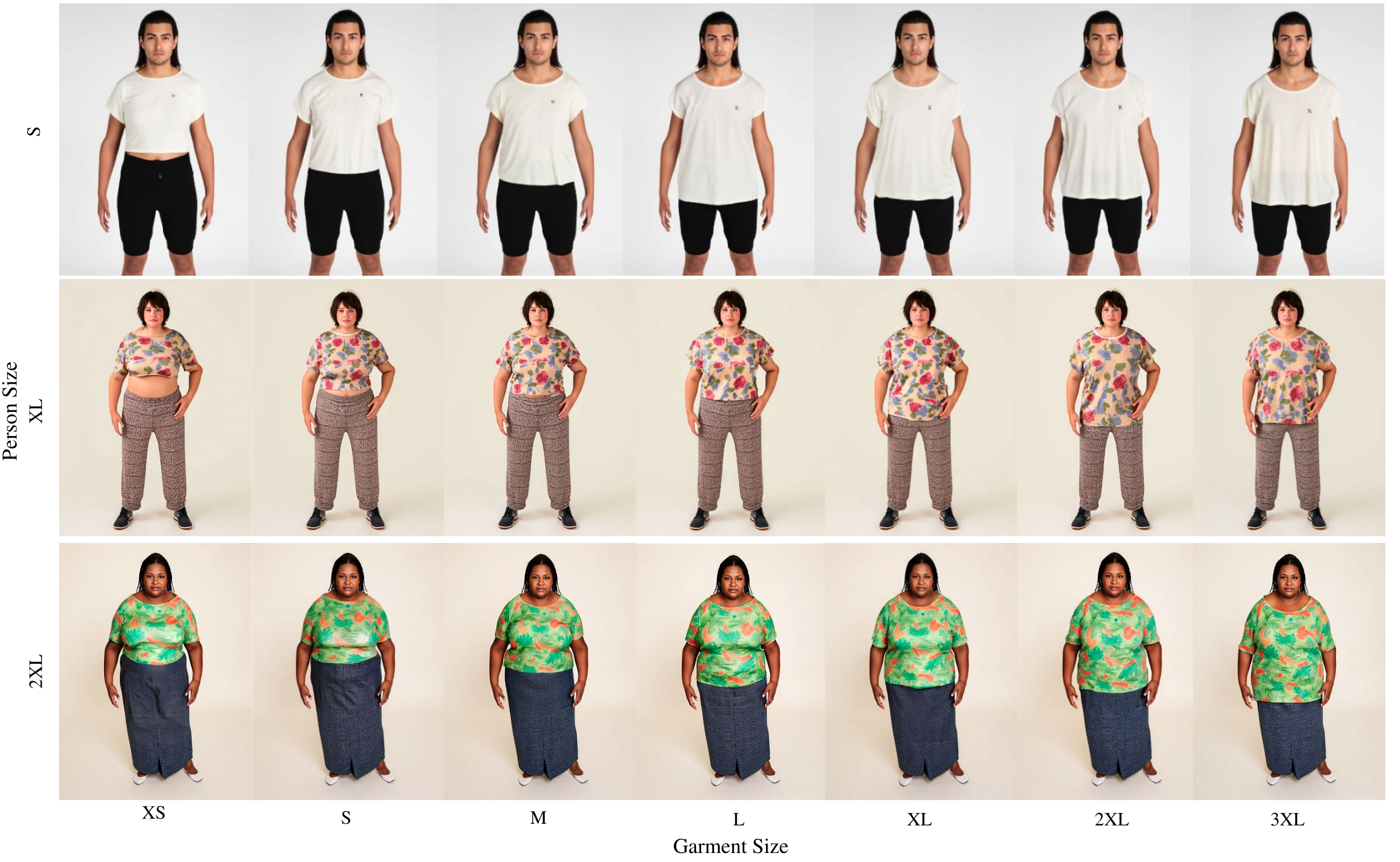}
    \vspace{-3mm}
  \Description{Independent size control.}
  \caption{Independent size control. Fit-VTO realistically visualizes garment fit across various sizes on a fixed person size.}
  \label{fig:size_matrix}
  \vspace{-5mm}
\end{figure*}

To evaluate Fit-VTO's ability to independently model person and garment size, we showcase the results of varying the garment size while keeping the person fixed in Figure~\ref{fig:size_matrix}. Fit-VTO realistically adjusts garment fit with respect to both garment and person sizes, while maintaining consistent person and garment appearance. In the appendix, we evaluate independent controllability of individual garment measurements and show that garment size controllability extends to images of real-world humans.

In Figure~\ref{fig:comparison}, we qualitatively compare our method to related works. Despite accurate texture warping, Any2AnyTryon~\cite{any2anytryon}, Nano Banana Pro~\cite{nanobananapro}, COTTON~\cite{chen2023size}, and IDM-VTON~\cite{idmvton} fail to accurately portray accurate garment fit according to the person and garment sizes. Nano Banana Pro, for example, produces aesthetically pleasing images, but lacks precise measurement grounding, leading to incorrect fit (e.g., overly loose or tight) relative to ground-truth (last column). COTTON also suffers from severe boundary artifacts due to errors in its pre-processing pipeline. In contrast, Fit-VTO respects person and garment measurements and visualizes accurate garment appearance. 

\subsection{Fit-VTO Quantitative Results}

We report quantitative metrics against related methods in Table~\ref{tab:comparisons}. Fit-VTO excels in nearly all VTO metrics on both real-world VITON-HD and synthetic FIT datasets. On VITON-HD, IDM-VTON's attains slightly stronger results, which is partially explained by its training directly on VITON-HD, whereas our base method (“ours”) is not. However, with additional VITON-HD finetuning, our method achieves comparable performance to IDM-VTON on VITON-HD. On the FIT dataset, Fit-VTO achieves superior size-aware IoU score, even compared to size-conditioned COTTON. These results indicate that our method effectively delivers high appearance fidelity, as well as incorporate size information for try-on.

\subsection{Ablations}
\begin{figure*}[!tbp]
  \centering  \includegraphics[width=0.85\linewidth]{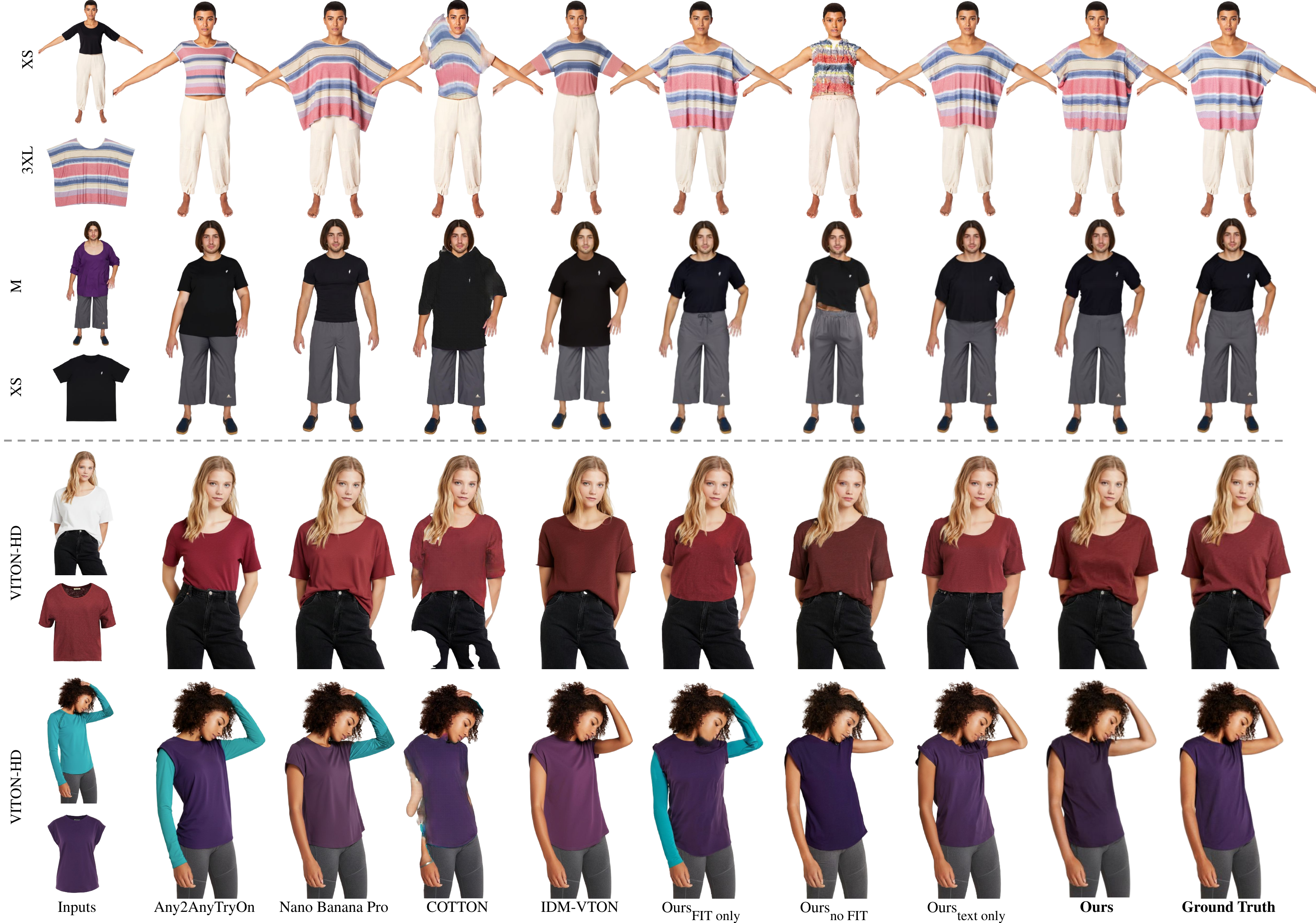}
  \vspace{-3mm}
  \Description{Qualitative Fit-VTO comparisons.}
  \caption{Qualitative comparisons. We compare Fit-VTO to related and ablated methods using synthetic FIT test data (top two rows) and real-world VITON-HD data (bottom two rows). Overall, our method best depicts the most accurate garment appearance and fit. 
  }
  \label{fig:comparison}
\end{figure*}

In our ablations, we evaluate the impact of our FIT dataset, measurements encoder, and real-world data supervision. As summarized in Table~\ref{tab:comparisons}, we compare (1) training without FIT data and only online fashion images (Ours$_{\text{no FIT}}$), (2) replacing our measurement encoder with pre-trained T5~\cite{t5} and CLIP~\cite{clip} text encoders used in the original Flux.1-dev~\cite{flux1dev} model (Ours$_{\text{text}}$), and (3) training with FIT data only (Ours$_{\text{FIT only}}$). See Figure~\ref{fig:comparison} for qualitative comparisons.

The FIT-only model performs well on FIT data, but degrades considerably on VITON-HD, as further evidenced in the bottom two rows of Figure~\ref{fig:comparison}. We attribute this to overfitting to the garments and poses FIT dataset, highlighting the importance of real-world training data for generalization. Conversely, the model trained without FIT data performs well on VITON-HD with respect to SSIM, FID, and LPIPS, but fails to model person-garment size relationships, as indicated by the significantly lower size-aware IoU. This demonstrates that real-world data with measurements predicted by VLM alone are insufficient for learning accurate fit. The text-only model performs moderately well on VITON-HD -- likely because it better preserves the pretrained knowledge from Flux.1-dev -- yet, this model fails to encode precise measurement information and exhibits a low IoU score. This indicates that pre-trained text encoders are not well-designed to represent structured numerical size inputs. Row 1 in Figure~\ref{fig:comparison} corroborates these findings: Ours$_{\text{no FIT}}$ and Ours$_{\text{text only}}$ exhibit significant errors in representing ill-fitting garment size, while our full method accurately show accurate garment fit.

Our full model achieves the best balance across both benchmarks, performing on par with the strongest variants on each domain while delivering high size-aware IoU on FIT. These results confirm that combining FIT supervision, real-world data, and our measurement encoder yield a model that is both robust to real imagery and sensitive to garment–person size relationships.
\section{Scope and Limitations} \label{sec:limitations}

Our work serves as a proof-of-concept demonstrating that synthetic data generation, grounded in physics-based simulation, is a promising way to overcome the scarcity of size-annotated data in virtual try-on. However, as an initial exploration, our current scope is intentionally constrained. We focus exclusively on upper-body garments in standardized front-facing views (full-body or cropped) and casual poses, thereby avoiding the complicated collision dynamics. Additionally, the structural diversity of our dataset is bounded by the capabilities of the GarmentCode engine, limiting our study to simple structural designs rather than complex, multi-layered apparel. Despite these constraints, our results validate the core hypothesis: that synthetic, physics-informed supervision can teach generative models to respect precise metric sizing. We believe this synthetic-to-real paradigm establishes a foundation for future research to scale up to complex, in-the-wild scenarios.

\begin{figure*}[!tbp]
  \centering  \includegraphics[width=\linewidth]{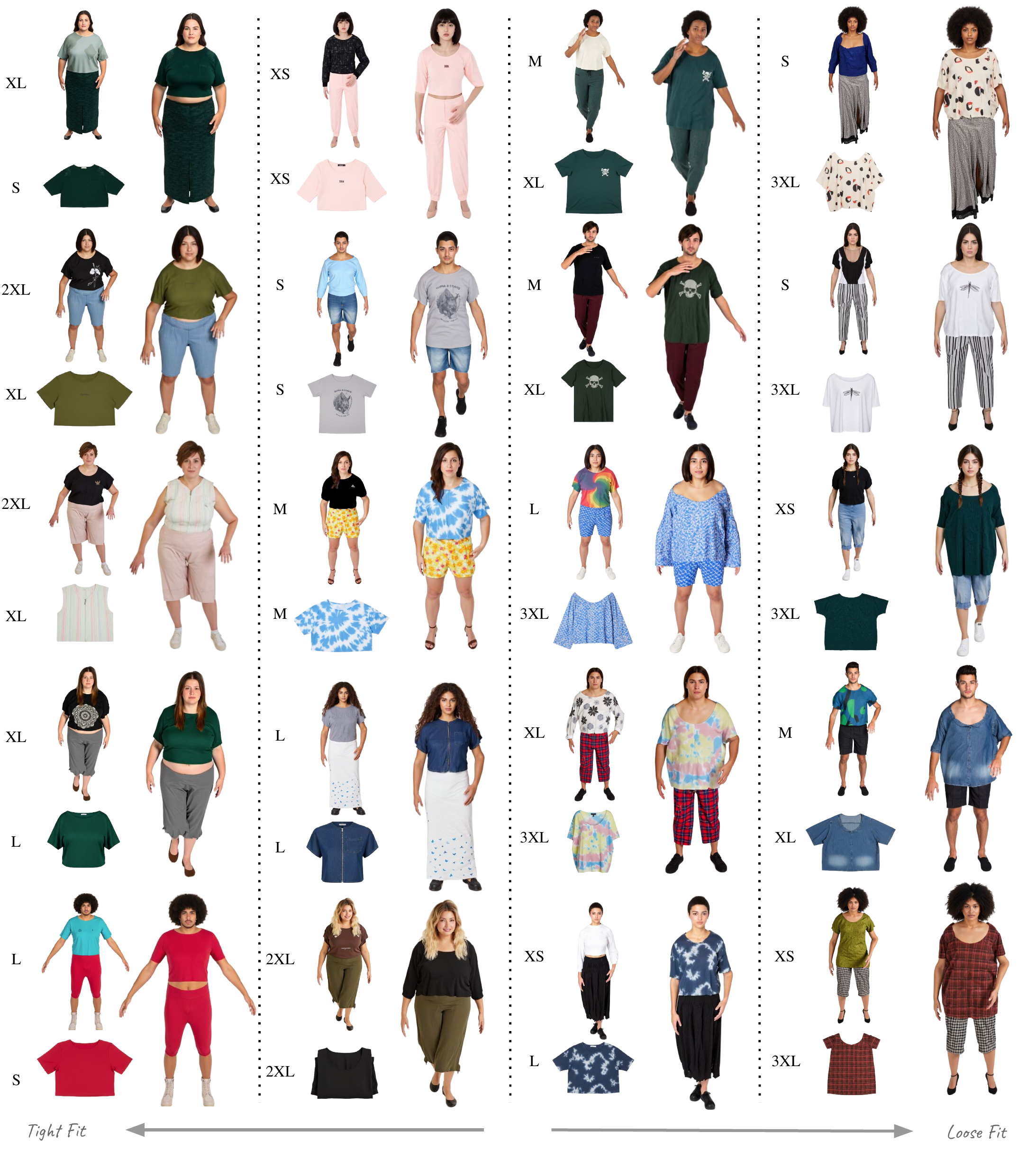}
  \Description{Qualitative Fit-VTO results.}
    \caption{Qualitative results. We show examples of Fit-VTO results on the synthetic FIT test dataset. Fit-VTO respects person and garment inputs, while also synthesizing realistic garment fit based on person and garment measurements (zoom in for details). For brevity, we approximate the full measurements with a size label (XS-3XL). See the appendix for our size categorization chart.}
  \label{fig:qualitative_vto}
  \vspace{-5mm}
\end{figure*}

We also identify two specific technical limitations.
First, accurately representing the degree of tightness in the data is challenging. While the feeling of wearing a tight garment or a very tight garment may vastly differ, the simulated appearance is almost identical -- fitted to the skin. As such, our dataset and VTO model do not represent varying degrees of tightness well (see left column of Figure~\ref{fig:qualitative_vto}). Furthermore, our Fit-VTO model is sensitive to correlations in measurements, limiting its ability to independently alter single measurements. For example, an increase in width frequently leads to a slight increase in length and sleeve length, as well. 
\section{Conclusions and Future Work}
\label{sec:discussion}

In this paper, we introduce \textbf{FIT}, the first large-scale dataset and benchmark for \textit{fit-aware} virtual try-on (VTO) consisting of over 1.13M samples. We also present \textbf{Fit-VTO}, a novel fit-aware VTO model designed to leverage FIT's rich person-garment size annotations. Across extensive comparisons to related and ablated methods, Fit-VTO demonstrates a clear advantage in modeling accurate garment fit, according to person and garment measurements.

\noindent \textbf{Future Work:} Our immediate next steps involve expanding the dataset scope beyond tops to include lower-body and full-body garments (e.g., pants, dresses). Additionally, while our current dataset supports basic pose variation, scaling up the diversity of poses and camera viewpoints remains a key objective to ensure robust performance across complex, real-world inputs.


\begin{acks}
We are grateful to the ARML team at Google for their valuable feedback and support during this project.
\end{acks}

\clearpage



\clearpage

\bibliographystyle{ACM-Reference-Format}
\bibliography{main}

\clearpage
\appendix 
\appendix
\setcounter{page}{1}

\twocolumn[ 
\begin{center}
    {\Huge \textbf{Appendix}}
    \vspace{1em}
\end{center}
]

\section{Additional FIT Dataset Details}
\label{sec:supp_dataset}

    In this section, we provide additional details and statistics about the FIT dataset. 
    
    \subsection{Size Categorization} \label{ssec:supp_size_categorization}
    In our figures, we frequently abbreviate the full person-garment measurements $[m_p, m_g]$ with coarse size labels for the person and garment such as XS, L, XL. We determine these size labels based on average measurement ranges as shown in Table~\ref{tab:size_categorization}. Note that the grouping of \textbf{size measurements} into \textbf{coarse size labels} (e.g. XS, L, XL) are only used for visualization and grouping purposes, not Fit-VTO training or evaluation.
    
    \subsection{Garment Fit Distribution} 
\begin{figure*}[!ht]
  \centering
  \includegraphics[width=\linewidth]{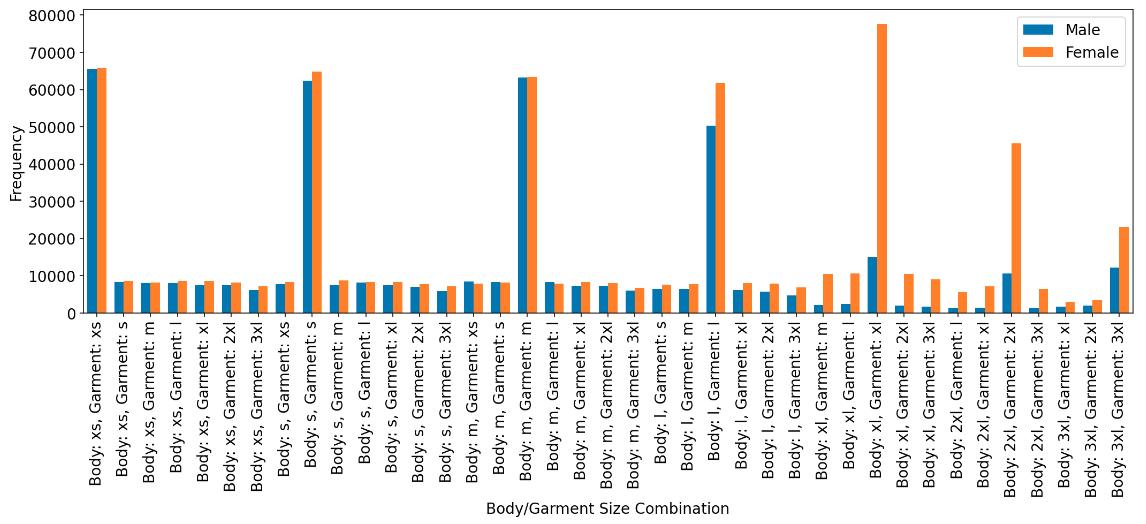}
  \caption{Garment fit distribution. We plot the frequency of each (body size, garment size) pairing in our dataset according to the size classification in Table~\ref{tab:size_categorization}.}
  \vspace{-5mm}
  \label{fig:fit_distribution}
\end{figure*}

    Our FIT dataset covers a diverse range of fit scenarios. In Figure~\ref{fig:fit_distribution}, we plot the distribution of person/garment size pairings as a histogram, showing that every reasonable fit scenario is represented, from very tight (e.g. size ``XL'' person wearing a size``M" garment) to very loose (e.g. size ``XS'' person wearing a size ``2XL" garment). Implausible fit pairings where the garment is more than 3 sizes smaller than the person (e.g. size ``3XL" person wearing a size``XS" garment) are not included.
    
    \subsection{Measurement Statistics} 
    In Table~\ref{tab:b_size_statistics} and Table~\ref{tab:g_size_statistics}, we report the minimum, mean, maximum, and standard deviations of the measurements for our body and garment meshes in the FIT dataset, respectively. Our dataset covers a wide range of body shapes and garment sizes.
    
    \begin{table}[htbp]
    \centering
    \small
    \caption{Body size statistics. We report the min, mean, max, and standard deviation of our garment measurements in cm. }
    \label{tab:b_size_statistics}
    \begin{tabular}{|c|c|c|}
        \hline
        & \textbf{Men's} & \textbf{Women's} \\
        Measurement & (min, mean, max, std) & (min, mean, max, std) \\
        \hline
        Bust & (87, 101, 141, 10) & (83, 100, 136, 13)  \\
        Height & (155, 174, 194, 8.0)	&  (151, 170, 196, 9.0) \\
        Hips & (88, 101, 125, 6.0) &  (89, 104, 127, 10)  \\
        Waist & (70, 86, 141, 13) &  (61, 85, 130, 17) \\
        \hline
    \end{tabular}
\end{table}
    \begin{table}[htbp]
    \centering
    \small
    \caption{Garment size statistics. We report the min, mean, max, and standard deviation of our garment measurements in cm. }
    \label{tab:g_size_statistics}
    \begin{tabular}{|c|c|c|}
        \hline
        & \textbf{Men's} & \textbf{Women's} \\
        Measurement & (min, mean, max, std) & (min, mean, max, std) \\
        \hline
        Width & (77, 112, 169, 16) & (75, 110, 169, 16)  \\
        Length & (29, 53, 76, 8.0)	&  (29, 51, 76, 7.5) \\
        Sleeve Length & (0.0, 30, 79, 17) &  (0, 29, 79, 17)  \\
        \hline
    \end{tabular}
\end{table}
    \begin{table*}[ht]
    \centering
    \small
    \caption{Body size categorization statistics. We report the min and max for each body measurements and size label in cm. }
    \label{tab:size_categorization}
    \begin{tabular}{|c|c|c|c|c|c|c|c|}
        \hline
        & \multicolumn{2}{c|}{Bust} & \multicolumn{2}{c|}{Waist} & \multicolumn{2}{c|}{Hips} \\
        \hline
        & \textbf{Men's} & \textbf{Women's}  & \textbf{Men's} & \textbf{Women's}  & \textbf{Men's} & \textbf{Women's} \\
        Size & (min, max) & (min, max) & (min, max) & (min, max) & (min, max) & (min, max) \\
        \hline
        XS & (86, 91) & (79, 84)  & (71, 76) & (58, 64)  & (91, 96) & (84, 89)\\
        S & (91, 96)&  (86, 89) & (76, 81) & (66, 67) & (96, 101)  &  (91, 94) \\
        M & (96, 101) &  (90, 95) & (81, 86) &  (71, 75) & (101, 106) &  (97, 102) \\
        L & (101, 106) &  (96, 104) & (86, 91) &  (86, 91) & (106, 111)  &  (106, 111)  \\
        XL & (106, 117) &  (105, 116) & (91, 103) &  (85, 97) & (111, 120) &  (112, 121)\\
        2XL & (111, 127) &  (112, 125) & (96, 115)	&  (91, 105) & (120, 134) &  (120, 130)\\
        3XL & (127, 147) &  (117, 135) &(115, 137)& (107, 127) & (134, 145) & (125, 137)\\
        \hline
    \end{tabular}
\end{table*}

    \begin{figure}[h]
  \centering
  \includegraphics[width=\linewidth]{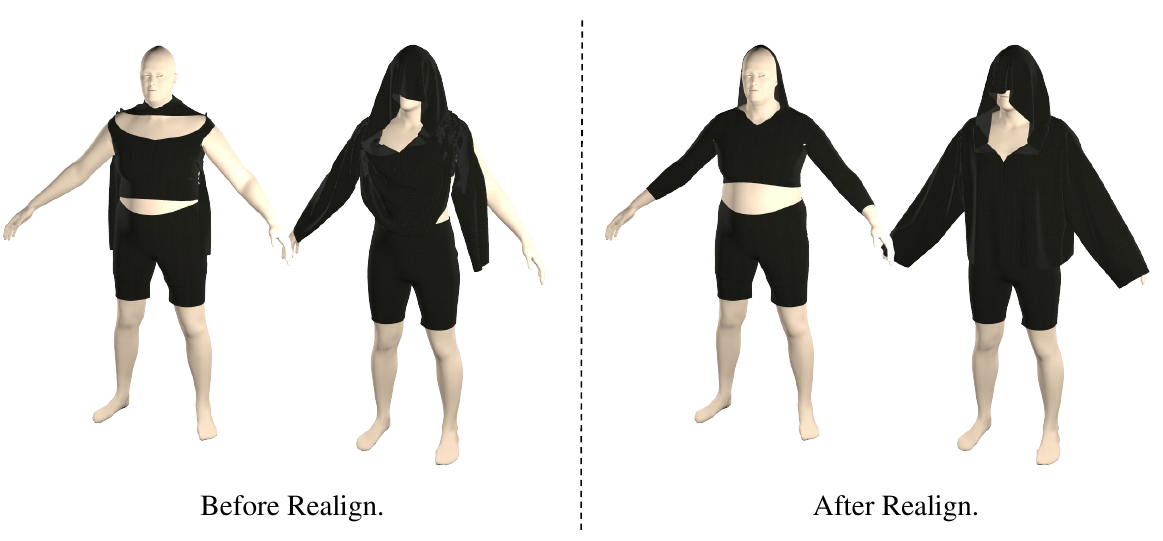}
  \caption{During cross-body draping, the initial boxmeshes are often misaligned with the target human models, causing draping failures (left). We explicitly realign the boxmesh to ensure successful simulation (right).}
  \label{fig:boxmesh_realign_supp}
\end{figure}

    \begin{figure*}[!tbp]
  \centering  \includegraphics[width=\linewidth]{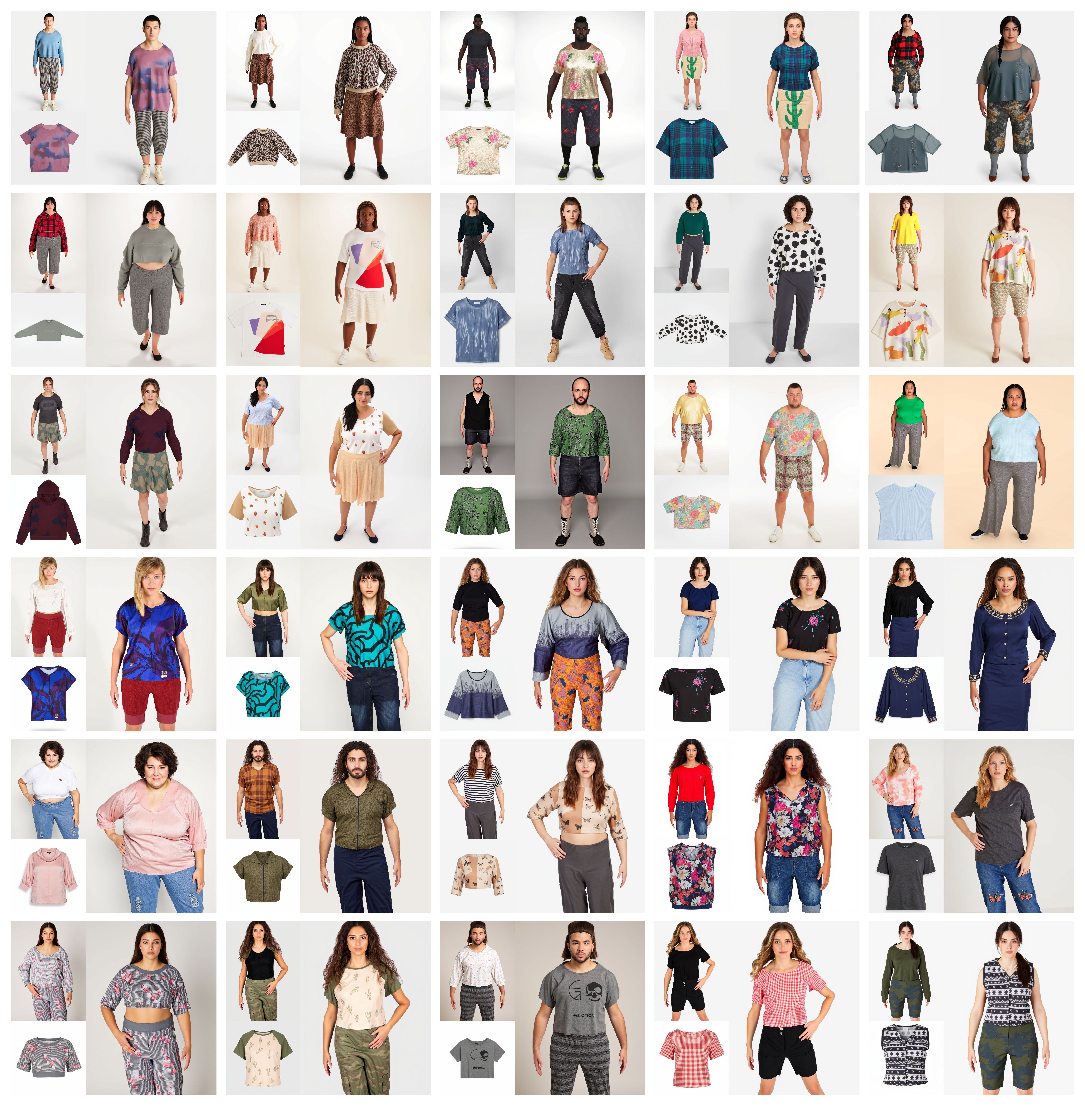}
  \caption{\textbf{Additional dataset examples.} In each example, we show the paired person image (top left), garment image (lower left), and the target try-on image (right).}
  \label{fig:additional_data}
\end{figure*}

    \subsection{Additional Dataset Examples}
    We present additional examples of try-on triplet data in our dataset in Figure~\ref{fig:additional_data}.

\section{Additional Details on Data Generation Pipeline}

\subsection{Cross Draping vs. Linear Size Change}
The standard GarmentCode pipeline computes sewing patterns based on a design template and target body parameters, yielding a 3D garment that is well-fitted to the wearer. To generate ill-fitting examples (e.g., oversized or undersized), a naive baseline would be to linearly scale the garment parameters. However, this fails to capture real-world sizing dynamics, as garment grading rules are non-linear and distinct from simple geometric scaling. To address this, we propose a cross-draping strategy. Instead of manipulating the mesh directly, we instantiate a separate ``source'' body in a different size, generate a pattern fitted to that body, and then drape the resulting garment onto the original target body. This process simulates the physical reality of a person wearing a garment designed for someone else, resulting in significantly more natural and realistic ill-fitting dynamics compared to simple linear scaling.

\subsection{Boxmesh Realignment}
Cross-draping a sewing pattern onto a target body mesh of a different size creates misalignments between the boxmesh panels and target body parts, which can lead to draping errors. We implement boxmesh realignment (Section 3.2 in main) as a critical step for successful cross-body draping. See Figure~\ref{fig:boxmesh_realign_supp} for a visual comparison with and without boxmesh alignment.

To align a given sewing pattern $p_{\text{tgt}}$(might be ill-fit to size $s_p$ body) with a target body mesh of size $s_p$, we use a different, well-fitted (i.e. generated on size $s_p$ body) sewing pattern $p_{\text{ref}}$ as a reference. We then align the panels of the $p_{\text{tgt}}$ to the spatial locations of $p_{\text{ref}}$. This ensures that $p_{\text{tgt}}$ is aligned to the target body mesh. Additionally, we observed that significant discrepancies between the human model's arm angle and the initialized sleeve panel angle can cause arm-sleeve penetrations. To mitigate this, we adjust the sleeve angle to match the arm angle prior to simulation.

\subsection{Retexturing Model Training Data}
We train our retexturing model on a dataset of 50k real-world person images. We use the person images in VITON-HD and additionally scraped online images featuring modeling posing in front of the camera with a studio background. We use Sapiens~\cite{khirodkar2024sapiens} to estimate normal and segmentation maps and Gemini~\cite{gemini} to generate prompts describing the garment textures and designs. We enforce a structured prompt format containing two sentences (one per garment piece). This facilitates inference for paired generation: since only the top garment is swapped, we update the first sentence corresponding to the top, while the second sentence remains frozen to preserve the bottom garment.

\subsection{Reposing}
Since GarmentCode simulation produces results exclusively in a static A-pose, we employ a customized reposing pipeline to repose the 3D simulated meshes, thereby expanding the dataset's diversity and improve model generalization. In total, we sample from 528 distinct target poses and repose each sample into a randomly chosen pose from the pool, prioritizing casual stances commonly encountered in real-world try-on scenarios.


\begin{figure*}[!tbp]
  \centering
  \includegraphics[width=0.95\linewidth]{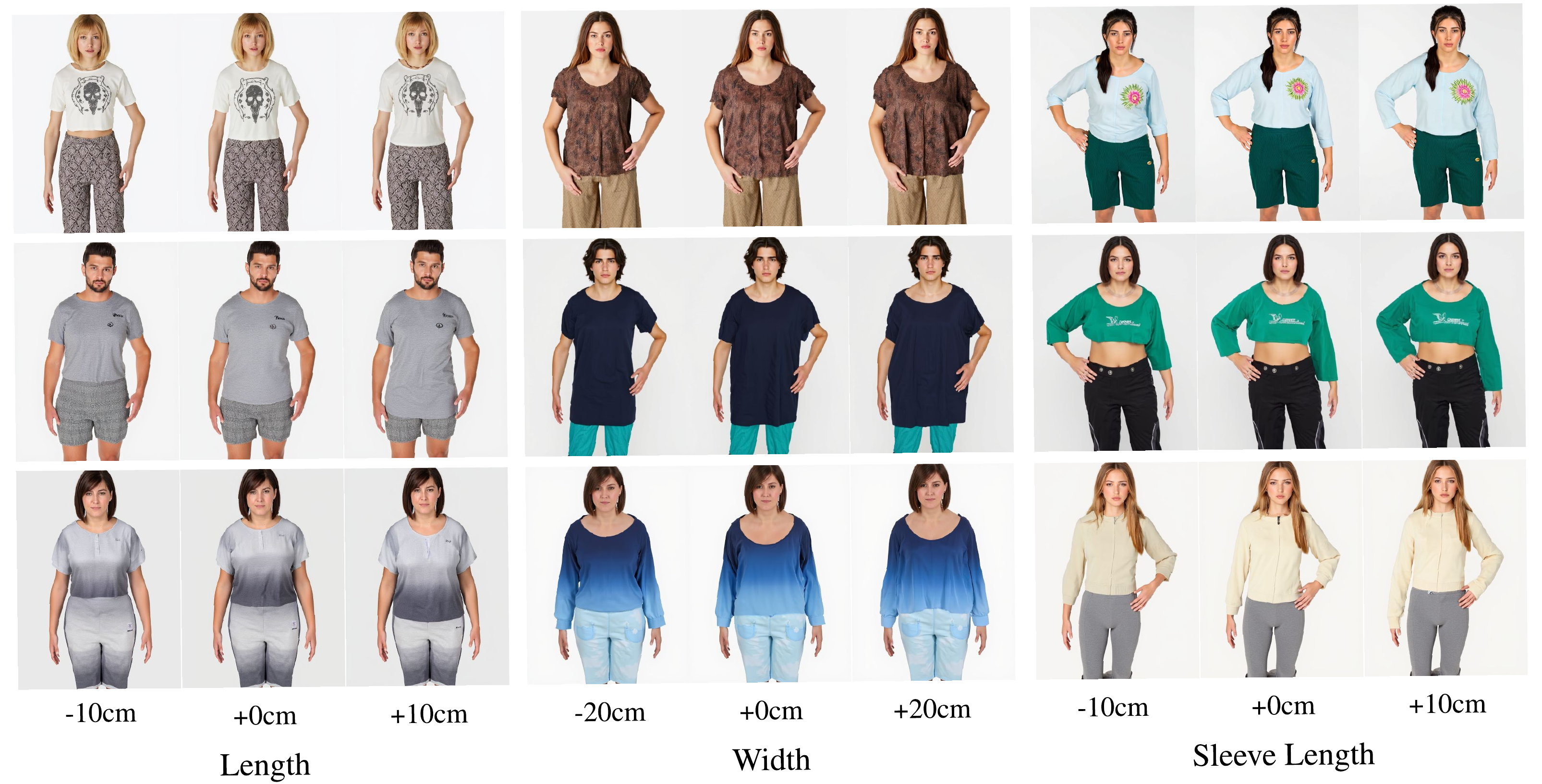}
    \vspace{-3mm}
  \Description{Qualitative resizing results.}
  \caption{Qualitative resizing results. Fit-VTO adapts garment fit according to individual garment measurements. We show results of independently shrinking and growing the length, width, and sleeve length with respect to the original value. Please zoom in for details.}
  \label{fig:resizing}
\end{figure*}

\section{Additional Resizing Results}
    We further evaluate independent controllability of individual garment measurements in Figure~\ref{fig:resizing}. Fit-VTO realistically adjusts specific garment dimensions with respect to measurement changes, while preserving the non-adjusted garment dimensions, as well as person and garment appearance.
    
\begin{figure*}[!tbp]
  \centering
  \vspace{3em}
  \includegraphics[width=\linewidth]{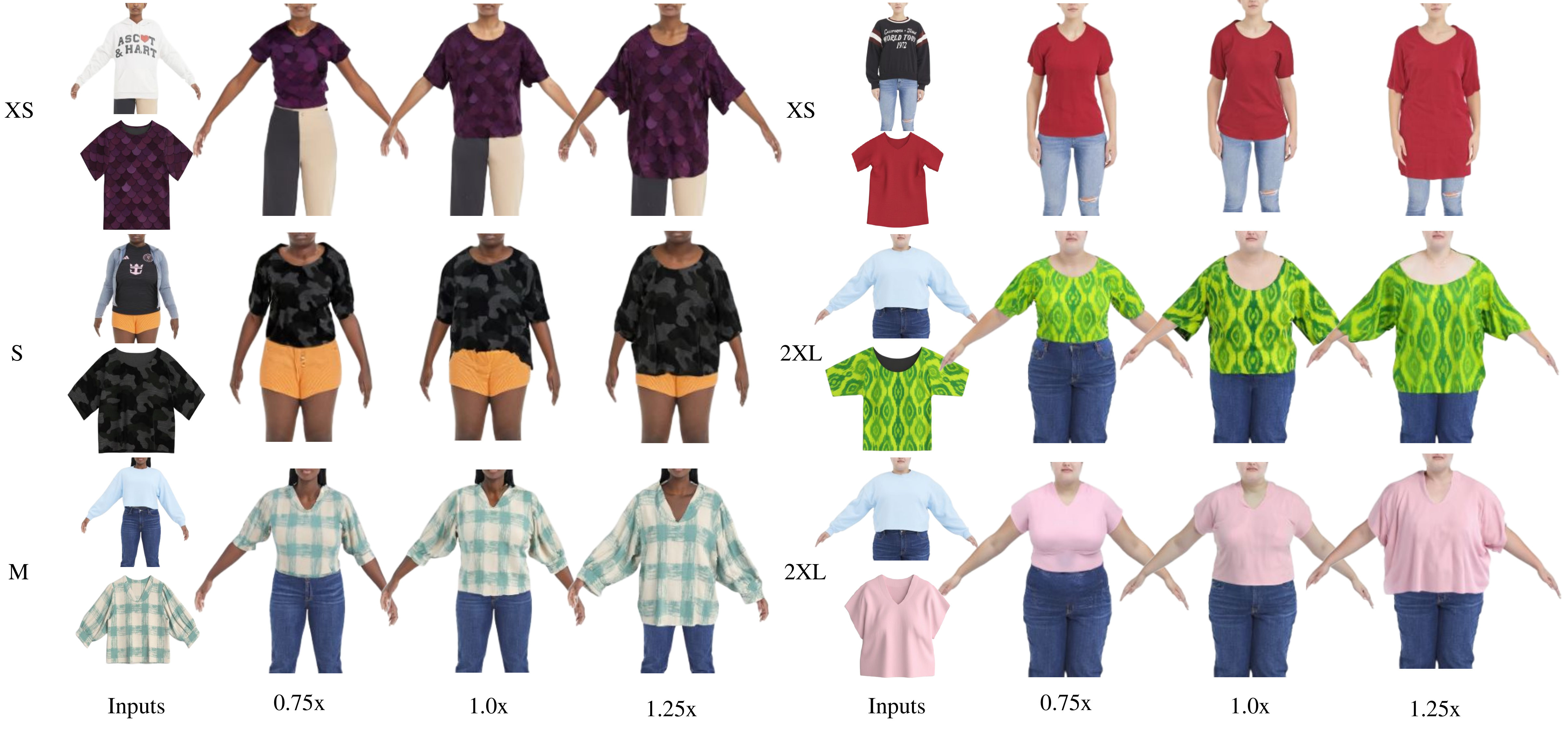}
  \caption{Real-world resizing results. We show Fit-VTO try-on performance on real-world person images using varying garment sizes. Fit-VTO realistically shrinks and grows the garment fit according to uniform adjustments to the garment measurements--length, width, and sleeve length--with respect to their original values (1.0x). The size label to the left of each example corresponds to the person's body size. Since real-world garment images with precise measurements are difficult to acquire, we use our synthetic garment images and measurements.}
  \label{fig:photoshoot-resizing}
\end{figure*}

    To show how Fit-VTO generalizes to real-world images, we provide additional resizing results on real-world person images in Figure~\ref{fig:photoshoot-resizing}. In these examples, the person and person measurements are captured from real human subjects, and the garment layflat image and measurements are randomly chosen from the FIT test dataset.

\section{Failure Cases}
\begin{figure*}[!tbp]
  \centering
  \includegraphics[width=\linewidth]{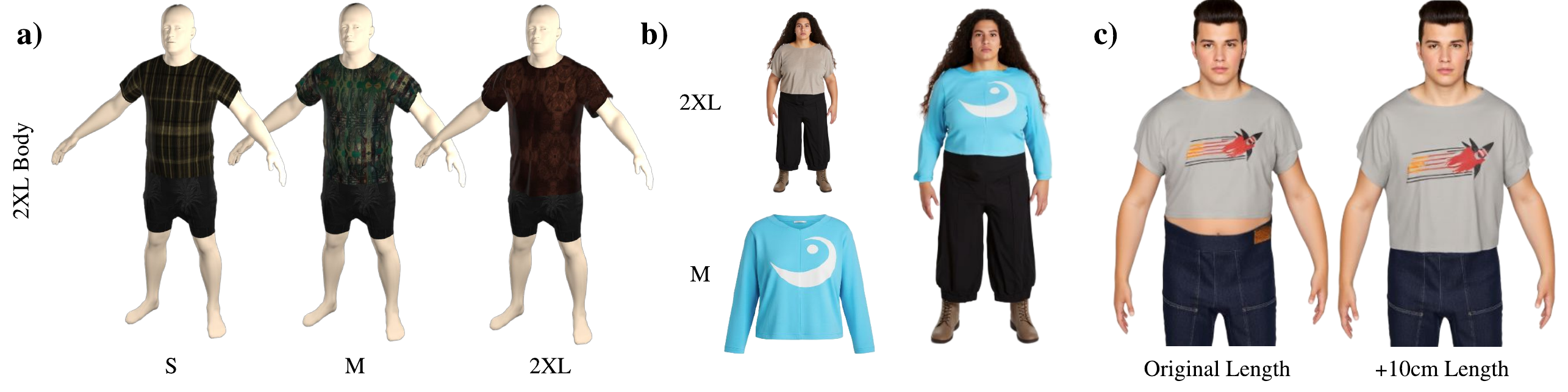}
  \caption{Failure cases. (a) GarmentCode~\cite{GarmentCode} simulation does not model varying degrees of tightness well, leading to similar-looking fit for all garment sizes smaller than the body size. (b) As a result of (a), it is difficult to tell the level of tightness in FIT try-on images. (c) Due to the correlations in measurements across sizes, adjustments to individual measurements may also lead to undesired changes in other dimensions. In this example, increasing garment length also increases garment width. }
  \label{fig:failure_cases}
\end{figure*}

    We show qualitative examples of the limitations of our method in Figure~\ref{fig:failure_cases}. These include a limited ability to represent varying degrees of garment tightness in GarmentCode~\cite{GarmentCode}, which we leave to future work. Another limitation is that garment measurements are often correlated in our FIT data (e.g. larger width correlates positively with larger length). As a result, with our Fit-VTO model, changing one measurement may lead to an unintentional change in another dimension.
    
\section{Comparisons to State-of-the-Art}\label{sec:supp-sota}
    \subsection{VITON-HD Preprocessing}

    Due to the lack of paired data, we generated pseudo paired-person images $I_p$ for every image in the VITON-HD~\cite{viton-hd} using Nano Banana Pro (see Section \ref{sec:vitonhd-prompt} for prompts). When running our Fit-VTO method, we set the each person and garment measurement to the null value (-1), same as the dropout value used during training.
    
    \subsection{Implementation Details}
        \noindent\textbf{Any2AnyTryon~\cite{any2anytryon}:}
        We used the model and code released from  the official implementation. For all evaluations, we used the ``dev\_lora\_any2any\_multi'' checkpoint. \\
    
        \noindent\textbf{COTTON~\cite{chen2023size}:}
        The officially released code and checkpoint trained on COTTON dataset was used. For evaluation on the VITON-HD test dataset, the default try-on mode was used. When running on FIT test dataset, the scaling parameter was computed as $r=\text{length} / \text{bust}$.\\
        
        \noindent\textbf{Nano Banana Pro~\cite{nanobananapro}:} For comparisons to  Nano Banana Pro~\cite{nanobananapro}, we input the paired-person image $I_p$, layflat garment image $I_g$, person-garment measurements $m$, and the prompt:

            \begin{quote}
                Edit $image_0$ so that the person wears garment in $image_1$ with size of person and garment described as \{measurement description\}.
            \end{quote}

        \noindent\textbf{IDM-VTON~\cite{idmvton}:}
        We used the model and code released from  the official implementation. For VITON-HD, the agnostic masks from the original data release were used. For FIT dataset, agnostic masks were computed from IDM-VTON preprocessing code. All hyper-parameters(e.g. number of diffusion steps) are set to be the recommended value from official release.

\section{LLM and VLM Prompts}\label{sec:supp_prompts}

In the follow sections, we provide the exact prompts used for all calls to LLM (Gemini~\cite{gemini}) and VLM (Nano Banana Pro~\cite{nanobananapro}) models in this paper.

\subsection{Head \& Shoes Generation}
\begin{quote}
    Change the head to make it look photorealistic. Add realistic $<$hair style$>$ hair, but the hair should always be behind the shoulder and never at the front. Add $<$shoe type$>$ if feet are visible. Make sure that everything else stays identical, including the human pose, garment shape, size, design and position.
\end{quote}

\subsection{Prompt Generation}
\begin{quote}
    Describe the garment in the image in two sentences. The first sentence should describe the top garment, and the second sentence should describe the bottom garment. Note that the input image is an illustration of the garment type, style and size - please ignore its existing texture. Please come up with some new description of the texture, logo and design. Add pocket, zipper, button, and other garment details if appropriate. Keep everything under 50 words.
\end{quote}

\subsection{Garment Try-Off}
\begin{quote}
    Create an in-shop product image of the top garment only against a plain white background.
\end{quote}

\subsection{Paired Person Image Generation}\label{sec:vitonhd-prompt}
\begin{quote}
    Generate a new image where the upper garment is changed, and keep everything else exactly the same, including the bottom garment, face, human pose, position etc.
\end{quote}

    

\subsection{Quality Assurance (QA)}

We leverage our LLM to filter out draping errors in $I_s$ that expose either person bust or groin area. We use the following two prompts to detect such errors:

    \begin{quote}
        Does the garment cover the person's chest? If so, return 'pass'. If not, return 'fail'.
    \end{quote}

    \begin{quote}
        Does the image contain a bottom garment (skirt, pants, underwear, boxers, leggings, or shorts) that covers the person's groin area?  If so, return 'pass'. If not, return 'fail'.
    \end{quote}

\section{Usage of LLM's}
\label{sec:llm-usage}

In addition to using an LLM~\cite{gemini} as described in Sections~\ref{sec:supp_dataset}, \ref{sec:supp-sota}, and \ref{sec:supp_prompts}, we leveraged an LLM to improve the grammar and clarity of our writing.




\end{document}